\pdfoutput=1

\documentclass[11pt]{article}

\usepackage{EMNLP2023}

\usepackage[utf8]{inputenc}

\usepackage{times}
\usepackage{latexsym}
\usepackage{multirow}
\usepackage{graphicx}
\usepackage{dingbat}
 \usepackage{makecell}
\usepackage{courier}
\usepackage{caption} 
\usepackage{tikz} 
\usepackage{xcolor}
\usepackage{booktabs}

\usepackage{arabtex}
\usepackage{utf8}
\setcode{utf8}
\usepackage{colortbl}
\usepackage{hyperref}

\usepackage[T1]{fontenc}
\usepackage[inline]{enumitem}
\usepackage{pifont}

\newcommand{\cmark}{\ding{51}}%
\newcommand{\xmark}{\ding{55}}%

\usepackage{microtype}

\usepackage{inconsolata}
\usepackage{soul}

\sethlcolor{mygreen} 
\newcommand*\rot{\rotatebox{90}}

\definecolor{mygreen}{rgb}{0.88, 1, 0.88}
\definecolor{myorange}{rgb}{1, 0.9, 0.8} 
\definecolor{myyellow}{rgb}{1, 1, 0.8} 
\definecolor{blue}{rgb}{0,0, 0.6}

\title{
Analyzing Multilingual Competency of LLMs in Multi-Turn Instruction Following: A Case Study of Arabic}

\author{Sabri Boughorbel \\
  Qatar Computing Research Institute \\
  Hamad Bin Khalifa University \\
  Doha, Qatar\\
  \texttt{sboughorbel@hbku.edu.qa} \\\And
  Majd Hawasly \\
  Qatar Computing Research Institute \\
  Hamad Bin Khalifa University \\
  Doha, Qatar\\
  \texttt{mhawasly@hbku.edu.qa} \\}

\begin{document}
\maketitle
\begin{abstract}
While significant progress has been made in benchmarking  Large Language Models (LLMs) across various tasks, there is a lack of comprehensive evaluation of their abilities in responding to multi-turn instructions in less-commonly tested languages like Arabic. Our paper offers a detailed examination of the proficiency of open LLMs in such scenarios in Arabic. Utilizing a customized Arabic translation of the MT-Bench benchmark suite, we employ GPT-4 as a uniform evaluator for both English and Arabic queries to assess and compare the performance of the LLMs on various open-ended tasks. Our findings reveal variations in model responses on different task categories, e.g., logic vs. literacy, when instructed in English or Arabic. We find that fine-tuned base models using multilingual and multi-turn datasets could be competitive to models trained from scratch on multilingual data. Finally, we hypothesize that an ensemble of small, open LLMs could perform competitively to proprietary LLMs on the benchmark.
\end{list}
\end{abstract}

\section{Introduction}
Recently, Large language models (LLMs) have brought about significant disruptions across various domains in both research and industry. LLMs have shown strong capability in solving and generalizing across diverse and complex tasks in natural language processing (NLP) and beyond. Moreover, their success in engaging in conversations and accurately following human instructions has been particularly noteworthy.  The recent surge in the availability of LLMs necessitates extensive benchmarking and evaluation.

\begin{figure}[!h]
    \centering
    \includegraphics[scale=0.43]{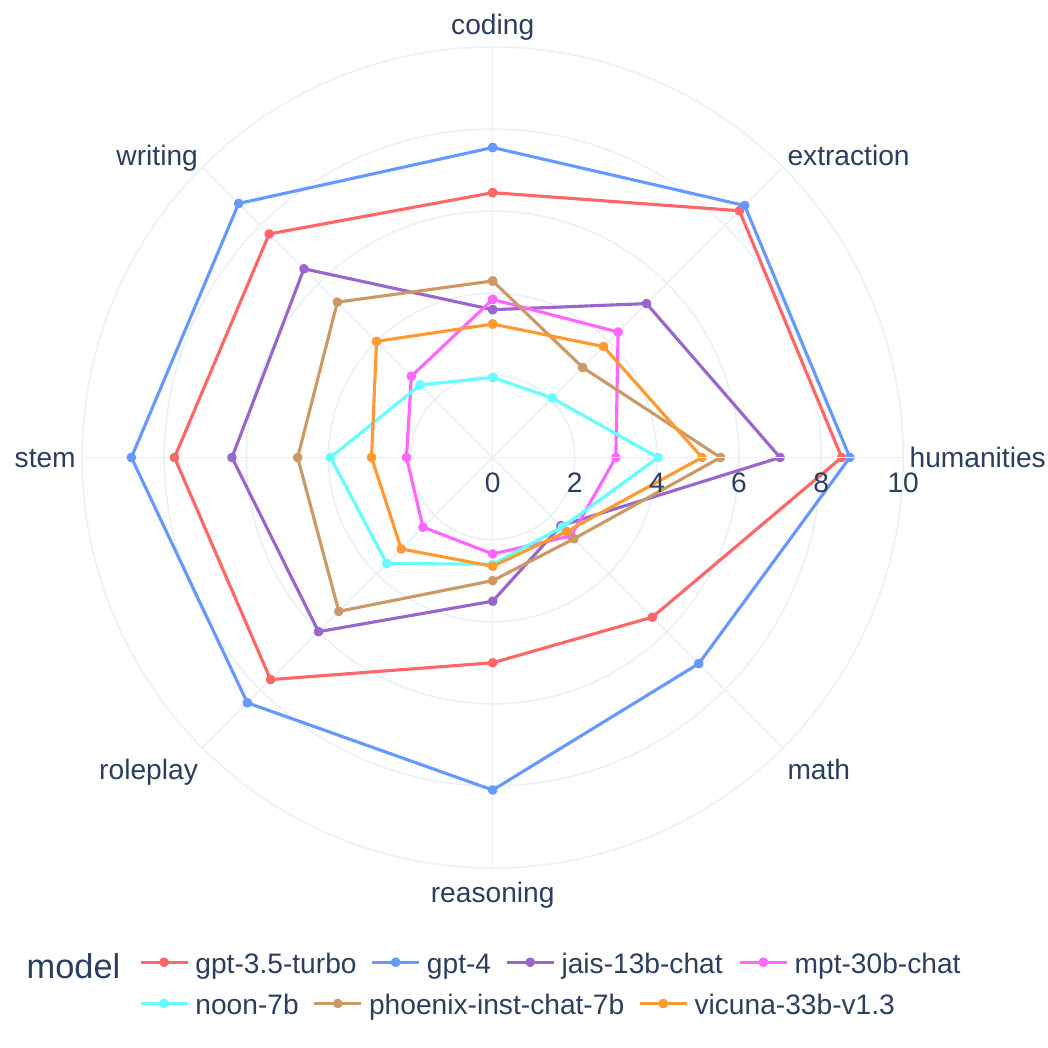}
    \caption{Performance scores per category for selected LLMs on the original MT-Bench~\cite{zheng2023judging} for English. The model responses are evaluated by GPT-4 and scored on a scale
of 1 to 10 using criteria of helpfulness, relevance, accuracy, depth, creativity, and level of detail.}
    \label{fig:radar_en}
\end{figure}


In this work, we analyze the competency of publicly-available, open LLMs when prompted with open-ended, multi-turn instructions in a language different than English. We compare the quality of these responses to the ones generated from equivalent instructions in English in order to identify the strengths and weaknesses of these models in terms of their multilinguality. Specifically, we study Arabic instructions, but the analysis could be repeated for any other language. Our study aim to answer the following questions:
\begin{itemize}[leftmargin=*]
    \item \textit{How do open LLMs fare in following open-ended instructions written in Arabic? and how do  they compare to GPT models?}
\item \textit{What is the effect of specifically targeting Arabic when training a model?}
\item \textit{What is the effect of specifically fine-tuning on Arabic multi-turn instructions?}
\item \textit{How to select a good starting point LLM model to fine-tune for Arabic instruction following? }
\end{itemize}

We start by a brief overview of the LLM benchmarking effort in Section~\ref{sec:benchmarking}. We introduce \textsc{Arabic MT-Bench} in Section~\ref{sec:Ar_mt_bench} as an analysis tool for multilingual instruction following. Then, we attempt to answer the proposed  questions through a number of analyses in Section~\ref{sec:results}. Finally, we conclude  in Section~\ref{sec:conclusion} with some insights and recommendations for pushing forward the competency of Arabic LLMs. 

\section{LLM Benchmarking}
\label{sec:benchmarking}
LLMs have shown capabilities that go far beyond traditional NLP tasks, such as text classification  or multi-choice question answering in some target natural language. Their ability to generate human-like text and engage in long conversations in any topic have opened up a multitude of novel opportunities and horizons that transcend tasks and languages. However, many existing benchmarks for LLMs are still anchored in the conventional NLP paradigm {or support English only}. Consequently, these benchmarks exhibit limitations when it comes to evaluating the proficiency of LLMs in open-ended generation, multi-turn tasks, or in  languages other than English.

\subsection{Conventional benchmarks}
Some of the recent effort in this category include projects such as HELM~\cite{liang2022holistic} 
and Evaluation Harness~\cite{eval_harness} which are platforms for LLM benchmarking. Also, standardized datasets such as MMLU~\cite{hendrycks2021measuring}, HellaSwag~\cite{zellers-etal-2019-hellaswag}, TruthfulQA~\cite{lin-etal-2022-truthfulqa}, ARC~\cite{mihaylov-etal-2018-suit} and OpenbookQA~\cite{clark2018think}, amongst many others, are used to evaluate core LLM capabilities such as commonsense reasoning, math, question answering, and factuality. {In addition, some recent works targeted Arabic language specifically with suites of tasks and datasets, e.g.~\cite{khondaker2023gptaraeval,abdelali2023benchmarking,alyafeai2023taqyim}}.

These benchmarks require specification of prompts per-task {and model}, in addition to post-processing functions  to validate model answers {against a gold standard}, which might not be straightforward and could prove time-consuming. Moreover,  with publicly available answer sets, there is always the potential risk of contamination to the training data of language models. Furthermore, some of these benchmarks have been shown to diverge in certain cases from human judgment~\cite{zheng2023judging}, 
possibly due to their narrow focus. 

\subsection{Instructional and conversational benchmarks}
Recent efforts on instruction-following benchmarks, such as Flan~\cite{longpre2023flan} and Super-NaturalInstructions~\cite{wang2022super}, or conversational benchmarks, such as OpenAssistant~\cite{kopf2023openassistant}, CoQA~\cite{reddy2019coqa} and MMDiag~\cite{feng2022mmdialog}, {present a more sophisticated and comprehensive challenge to LLMs, but they} are mostly limited to English, and the diversity of the questions are insufficient for the most advanced LLMs. Translating such datasets to other language is not a straightforward task, as it requires a large effort to manually curate the translated questions and answers for the purpose of ensuring high quality in the target language.

\subsection{Evaluating open-ended questions}
When it comes to open-ended tasks, such as creative writing, human evaluation of LLM responses is indispensable. Here, a human-in-the-loop acts as a judge to directly score an LLM response or to rank responses of multiple LLMs for the best answer on some question. However, achieving a reliable benchmark this way  is resource-intensive and lacks scalability. In one application, LMSYS Chatbot Arena\footnote{{\url{https://chat.lmsys.org}}}, which is a crowd-sourced LLM evaluation platform, allows users to use freestyle prompts for two randomly-selected LLMs before voting for the better response. Benchmarking using this approach,  {while very powerful,} is challenging as it compares  models evaluated on different prompts.

\begin{table}[t]
\begin{tabular}{p{0.02\columnwidth}|p{0.04\columnwidth}|p{0.75\columnwidth}}
                  \hline
\multirow{7}{*}{\rot{Writing}}& \multirow{4}{*}{T1} & Craft an intriguing opening paragraph for a fictional short story. The story should involve a character who wakes up one morning to find that they can time travel.    \\ \cline{2-3}
                          & \multirow{3}{*}{T2} &Summarize the story with three bullet points using only nouns and adjectives, without verbs.   \\ \Xhline{2\arrayrulewidth}

\multirow{7}{*}{\rot{Reasoning}}& \multirow{3}{*}{T1} & David has three sisters. Each of them has one brother. How many brothers does David have? \\ \cline{2-3}
                          & \multirow{4}{*}{T2} & If we change the previous question and assume that each sister of David has two brothers, how many brothers would David have? \\ \Xhline{2\arrayrulewidth}

\multirow{5}{*}{\rot{Math}}     & \multirow{3}{*}{T1} & The vertices of a triangle are at points (0, 0), (-1, 1), and (3, 3). What is the area of the triangle?   \\ \cline{2-3}
                            & \multirow{2}{*}{T2} & What's area of the circle circumscribing the triangle?     \\ \hline
\end{tabular}
\caption{A sample of questions from MT-Bench in categories Writing, Reasoning and Math. T1 and T2 denote the first turn and second turn (follow-up) questions, respectively.}
\label{tbl:examples}
\end{table}

An alternative approach that has recently emerged is the employment of an LLM to act as a judge of the responses of other LLMs. 
MT-Bench (Multi-Turn Benchmark)~\cite{zheng2023judging} utilizes this approach on a standard set of 80 {open-ended} questions of eight categories; namely: writing, extraction, reasoning, math, coding, role-play, humanities, and STEM. Moreover, it assesses the ability of an LLM to maintain a conversation by asking it a follow-up question that is based on its response to the first question. 
Examples of the MT-Bench questions are shown in Table~\ref{tbl:examples}.  These examples illustrate the level of open endedness and complexity  of the questions, and the dependency of the follow-up question on the first turn.

MT-Bench prompts a judge LLM with an instruction to rate the responses on a scale of 1-10 (where 1 indicates failure in answering the question and 10 indicates a perfect answer), clearly defining the evaluation task and criteria. Also, the judge LLM is asked to provide an explanation for the suggested score. This approach has been shown to have an agreement rate of 85\% with human evaluation when GPT-4 is used as a judge~\cite{zheng2023judging}, which was also  found to be higher than human-human agreement (81\%). {MT-Bench scores for selected LLMs are shown in Figure~\ref{fig:radar_en}}.

The approach of MT-Bench is versatile and scalable {as it delegates the resource-intensive scoring of open-ended questions to the judge LLM.} Moreover,  it could be extended to benchmarking LLMs in other languages by translating  the  benchmark dataset to the target language as long as a good judge LLM exists for that language. For Arabic, GPT-4 is highly-competent and has showed a good level of {proficiency~\cite{khondaker2023gptaraeval,abdelali2023benchmarking,alyafeai2023taqyim}}. Therefore, it is eligible to be used as a judge for Arabic responses. Moreover, by using the same  prompt for judging English and Arabic responses for the original and translated versions of the  same question, it is even possible to contrast the multilingual skills of an LLMs at a question and a category level.

\section{\textsc{Arabic MT-Bench}}
\label{sec:Ar_mt_bench}
In this work, we develop an Arabic version of MT-Bench. First, we  auto-translated the original benchmarking questions using Google Translate. A thorough manual curation of the translations is then performed. This step is essential to ensure the quality of the question set and hence the responses and the judgment. 
For example, all people names in the questions were changed to Arabic names, and questions about correcting English grammatical errors were re-written. {See Table~\ref{tbl:example_curated} in Appendix~\ref{sec:appx_questions} for a sample of curated translated questions }\footnote{{\textsc{Arabic MT-Bench} is available at\\ \url{https://huggingface.co/spaces/QCRI/mt-bench-ar}}}.

In addition to the questions, the benchmark provides reference answers for reasoning, math and code questions that are passed to the LLM judge to aid in the judgment. {One option to get these reference answers in Arabic is to prompt GPT-4 with the translated Arabic questions directly, but we decided instead to translate the original reference answers from English to ensure that the Arabic scores for these three categories stay as close as possible to the English MT-bench scores.}

Finally, our initial evaluation showed that some LLMs tend to respond in English despite the question being in Arabic. Hence, we decided to add at the end of each question a clear instruction to the LLM to respond in Arabic (\<الرجاء الإجابة باللغة العربية>). We observed that, without having to modify the original judgment prompt, GPT-4, acting as an Arabic judge, has taken into consideration that instruction and scored lower responses in English. 

{Table~\ref{tbl:stats} gives an overview of the \textsc{Arabic MT-Bench} dataset.}

\begin{table}[h]
    \centering
    \begin{tabular}{l|c}
    \hline
        \small Number of question categories &\small 8 \\ \hline
        \small Number of questions per category &\small 10 \\ \hline
        \small Number of turns per question &\small 2 \\ \hline
        \small Number of reference answers &\small 30 \\ \hline
        
    \end{tabular}
    \caption{Statistics of \textsc{Arabic MT-Bench} dataset }
    \label{tbl:stats}
\end{table}

\begin{table*}[h!]
    \centering
    \small
    \begin{tabular}{c|p{0.85\textwidth}}
    \hline 
    \textbf{Rating} & \textbf{Justification summary} \\
    \hline
       \multirow{4}{*}{2} & Common issues in AI assistant responses include: not addressing user's question, providing irrelevant or repetitive information, lacking depth, creativity, and accuracy, not following user's specific instructions, and not using the requested language. Users often seek detailed, accurate, and creative answers tailored to their requests, but AI assistants sometimes fail to deliver, resulting in unhelpful or unsatisfactory responses. \\
       \hline
        \multirow{6}{*}{4} & Common issues in the AI assistant's responses include lack of depth, inaccuracies, language inconsistencies, and not directly addressing the user's question. Some responses are repetitive and do not provide comprehensive analysis or examples. To improve, the AI assistant should focus on directly answering the user's question, providing clear and accurate examples, maintaining language consistency, and offering detailed and informative explanations. Additionally, adhering to specific user instructions and avoiding repetition will enhance the overall quality of the responses. \\
        \hline
        \multirow{6}{*}{8} & AI assistants provide relevant, creative, and accurate responses to various user requests, demonstrating a good understanding of topics and user instructions. They offer helpful suggestions, clear explanations, and maintain requested languages. Responses cover a wide range of subjects, including summarization, problem-solving, and engaging in fictional conversations. However, there are occasional minor mistakes and areas for improvement in clarity and depth. Overall, AI assistants successfully address user questions, providing satisfactory and informative answers.\\
\hline    \end{tabular}
    \caption{Summaries provided by GPT-4 of the collection of judgment justifications for questiones rated 2, 4 and 8 across all models and tasks. This indicates some level of internal consistency of the \textsc{Arabic MT-Bench} scores.}
    \label{tab:rating_justification}
\end{table*}

\subsection{Score consistency}
{In order to answer the question: \textit{are the scores of \textsc{Arabic MT-Bench} consistent and coherent such that it could be used as a metric?}
} and to qualitatively assess the effectiveness of \textsc{Arabic MT-Bench}, we clustered the judgments across all models and categories by their numerical ratings,  then asked GPT-4 to summarize its justification texts for every score (1 to 10). In Table~\ref{tab:rating_justification} are examples of the justification summaries for some ratings.

\begin{table*}[ht]
    \centering
    \small
\begin{tabular}{l|c|r|l|c}
\hline
\textbf{Model} & \textbf{Base model} & \textbf{Size} &  \textbf{Training language}  & \textbf{Multi-turn}\\
\hline
\rowcolor[gray]{.93}\textit{GPT-4} & \_ & $>$175B & Multilingual & \cmark \\
\hline
\rowcolor[gray]{.93}\textit{GPT-3.5-turbo} & \_ & 175B & Multilingual & \cmark \\
\Xhline{2\arrayrulewidth}
\textit{Jais-13B-chat} & Jais-13B & 13B & EN, AR& \cmark\\
\hline

\textit{PolyLM-13B} & \_ & 13B & Multilingual & \xmark\\
\hline
\textit{MPT-30B-chat}  & MPT-30B & 30B & Primarily English & \cmark\\

\Xhline{2\arrayrulewidth}
\textit{LLaMa-2-13B-chat} &LLaMa-2-13B & 13B & Primarily English & \cmark\\
\hline
\textit{Tulu-30B} & LLaMa & 33B & Primarily English & \xmark\\
\hline
\textit{Guanaco-33B} & LLaMa & 33B & Primarily English & \xmark\\
\hline
\textit{Vicuna-33B-v1.3} & LLaMa & 33B & Primarily English  & \cmark\\
\Xhline{2\arrayrulewidth}

\textit{BLOOMZ-7B1 } & \_ & 7.1B & Multilingual & \xmark\\
\hline
\textit{BLOOMZ-7B1-MT} &BLOOMZ-7B1 & 7.1B & Multilingual & \xmark\\
\hline
\textit{Noon-7B} & BLOOM & 7B & Multilingual,  AR fine-tuning& \xmark \\
\hline
\textit{Phoenix-chat-7B} & BLOOMZ-7B1-MT & 7B & Multilingual  & \cmark\\
\hline
\textit{Phoenix-inst-chat-7B} & BLOOMZ-7B1-MT & 7B & Multilingual & \cmark\\
\hline

\end{tabular}
    \caption{Attributes of the chosen models for this study. \_ for the `Base model' indicates a model that has been trained from scratch. `Size' is in the number of parameters. `Training language' is the natural language/s that made up the pre-training and instruction datasets for the model, and `Multi-turn' refers to chat fine-tuning.}
    \label{tab:models}
\end{table*}

While qualitative, we could conclude from this analysis that the justifications are reasonably consistent across models and categories, indicating an acceptable level of impartiality. In addition to that, the correlation between scores using the Arabic and English benchmarks for strong models, as will be seen Section~\ref{sec:results}, is another supporting evidence for the viability of \textsc{Arabic MT-Bench} as a metric.

\section{Results and Discussion}
\label{sec:results}
\subsection{Model selection}
In addition to OpenAI GPT-3.5-turbo and GPT-4, which are only considered in this work to set an upper bound, a number of open LLMs  have been chosen for this study. Through preliminary evaluations on HuggingFace playground, some LLMs exhibited knowledge of Arabic despite not being purposefully trained for it. 
The criteria we adopted  for choosing  models involve:
\begin{itemize}[leftmargin=*]
\item the model is open-source. Some competitive proprietary models are not accessible to us.
\item the model size is 33B or less, a decision driven by constraints in hardware infrastructure.
\item the model is known to do well on the English benchmarks on the LMSYS   leaderboard\footnote{\url{https://chat.lmsys.org/?arena}} 

\end{itemize}
An overview of the chosen models can be seen in Table~\ref{tab:models}, and more details can be found in Appendix~\ref{sec:chosen_models}.
 

\subsection{How do open LLMs fare in following open-ended instructions written in Arabic?}

Table~\ref{tab:leaderboard} shows the model ranking based on the \textsc{Arabic MT-Bench} scores. The first, second and third columns of the tables give the model's average score for the first turn across all questions, the average score for the second turn across all questions, and the average of both, respectively. Per-category scores could be seen in Figure~\ref{fig:radar_ar}. For comparison, Figure~\ref{fig:radar_en} (and Table~\ref{tab:leaderboard_en} in Appendix~\ref{appx:results}) give the per-category scores for the original English MT-Bench for the same models.

\begin{figure}[!h]
    \centering
    \includegraphics[scale=0.43]{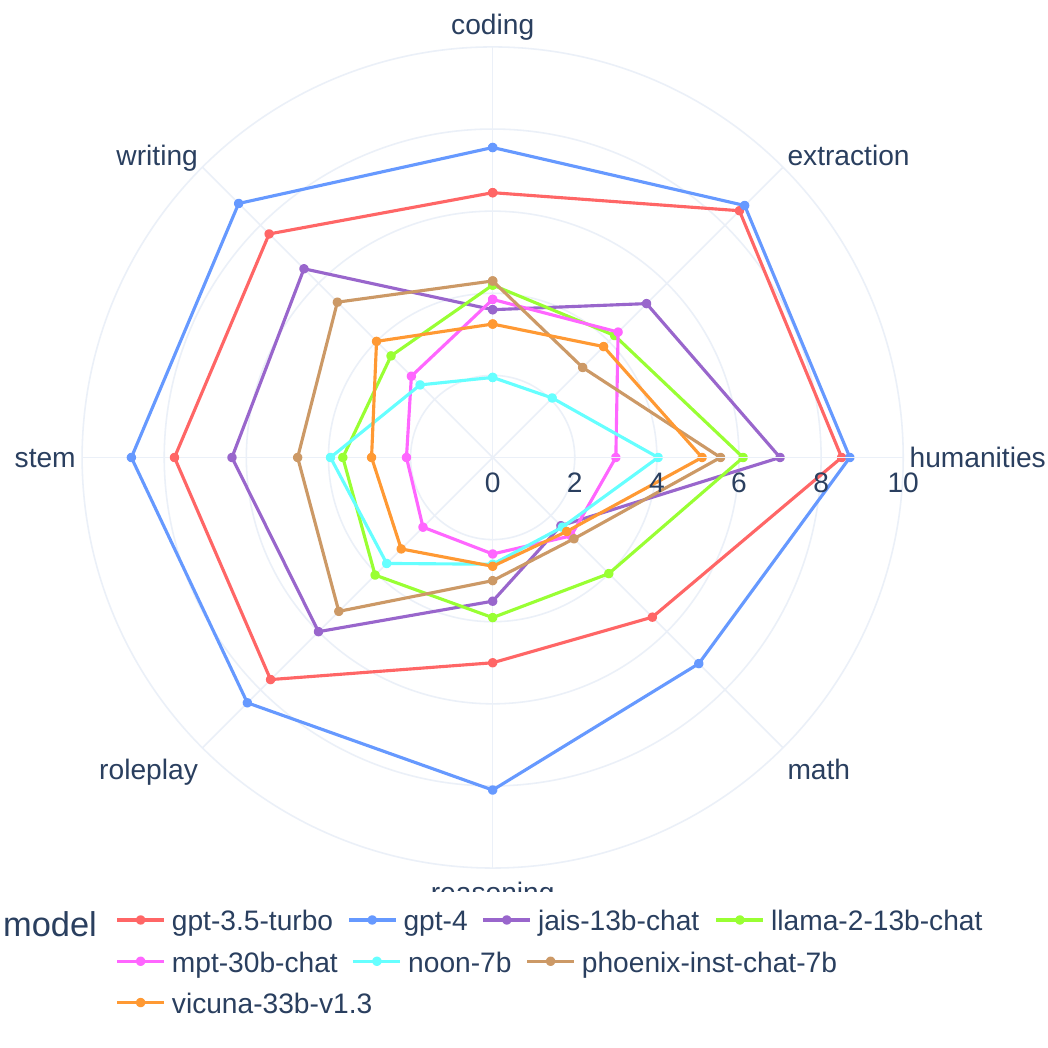}
    \caption{Performance scores per category for selected LLMs on our Arabic multi-turn benchmark. The model responses are evaluated by GPT-4 and scored on a scale of 1 to 10 using criteria of helpfulness, relevance, accuracy, depth, creativity, and level of detail.}
    \label{fig:radar_ar}
\end{figure}

\begin{table}[!t]
    \centering
    
\begin{tabular}{l|l|l|r}
\toprule
\textbf{Model} & \textbf{Turn1} &  \textbf{Turn2} & \textbf{Avg}\\
\midrule
\rowcolor[gray]{.93}\textit{GPT-4} & 8.41 & 8.12 & 8.27 \\
\rowcolor[gray]{.93}\textit{GPT-3.5-turbo} & 7.48 & 6.79 & 7.13 \\
\hline
\textit{Jais-13B-chat} & 5.01 & 5.14 & 5.08 \\
\textit{Phoenix-inst-chat-7B} & 4.84 & 3.70 & 4.27 \\
\textit{Llama-2-13B-chat} & 4.54& 3.86& 4.20\\
\textit{Phoenix-chat-7B} & 4.16 & 3.84 & 4.00 \\
\textit{Vicuna-33B-v1.3} & 3.44 & 3.43 & 3.43 \\
\textit{MPT-30B-chat} & 3.26 & 2.62 & 2.94 \\
\textit{Noon-7B} & 3.39 & 2.39 & 2.89 \\
\textit{Guanaco-33B} & 2.68 & 2.52 & 2.60 \\
\textit{PolyLM-13B} & 1.91 & 2.08 & 1.99 \\
\textit{Bloomz-7B1-mt} & 1.54 & 1.75 & 1.64 \\
\textit{Bloomz-7B1} & 1.29 & 1.54 & 1.41 \\
\textit{Tulu-30B} & 1.10 & 1.35 & 1.23 \\
\bottomrule
\end{tabular}
    \caption{Results of benchmarked LLMs on \textsc{Arabic MT-Bench} (scores between 1-10). showing for each model average scores per turn,  and average score across all questions and turns.}
    \label{tab:leaderboard}
\end{table}

As the results show, GPT-4 and GPT-3.5-turbo are better than any open LLM we tested by a large margin with average scores of 8.27 and 7.13 out of 10, respectively. Because GPT-4 is used as the judge, there exists the potential for bias in favor of its own responses, which  has been discussed in the MT-Bench paper~\cite{zheng2023judging}.

In the English MT-Bench, the two GPT models score 8.99 and 7.0, respectively.  Hence, GPT-4 is approximately one point lower in terms of the Arabic score compared to the English benchmark. By  manual inspection of the responses, we qualitatively confirm that the proficiency of GPT models in Arabic is lower than English as indicated by the scores
. Therefore, we compare the scores across Arabic and English benchmarks in Section~\ref{ssec:arabic_training}.

Overall, LLMs fine-tuned specifically for Arabic or for multilingual capabilities {(e.g. Jais, Phoenix)} are better than generic models such as some members of  the Llama family (e.g. Vicuna, Guanaco) in Arabic instruction following, even when smaller in size. The fine-tuning data and recipe matters significantly; 
for example, Phoenix-inst-chat-7B is much better then its predecessor Bloomz-7B1 or Bloomz-7B1-mt.

Jais-13B-chat is the best open model in Arabic in our evaluation. It achieves an average score of {5.08} out 10. The model has targeted Arabic and English in both pre-training and fine-tuning. Despite this, its relatively small size hinders it from being competitive with the best models. Also, it is still far on the English MT-Bench leaderboard from  models of comparable size, where the best model within 13B size in the English MT-Bench achieves a score above 6 out of 10 (see a selection of these scores in Table~\ref{tab:leaderboard_en} in the Appendix). Also, Jais-13B-chat model has the largest drop in performance in the second-turn questions on the English benchmark. Jais-13B-chat has been benchmarked internally using a similar approach to ours on private data accordingly to its technical report~\cite{sengupta2023jais}.

\begin{figure}[ht]
    \centering
    \includegraphics[trim=1cm 0.8cm 1cm 0.5cm,clip,scale=0.42]{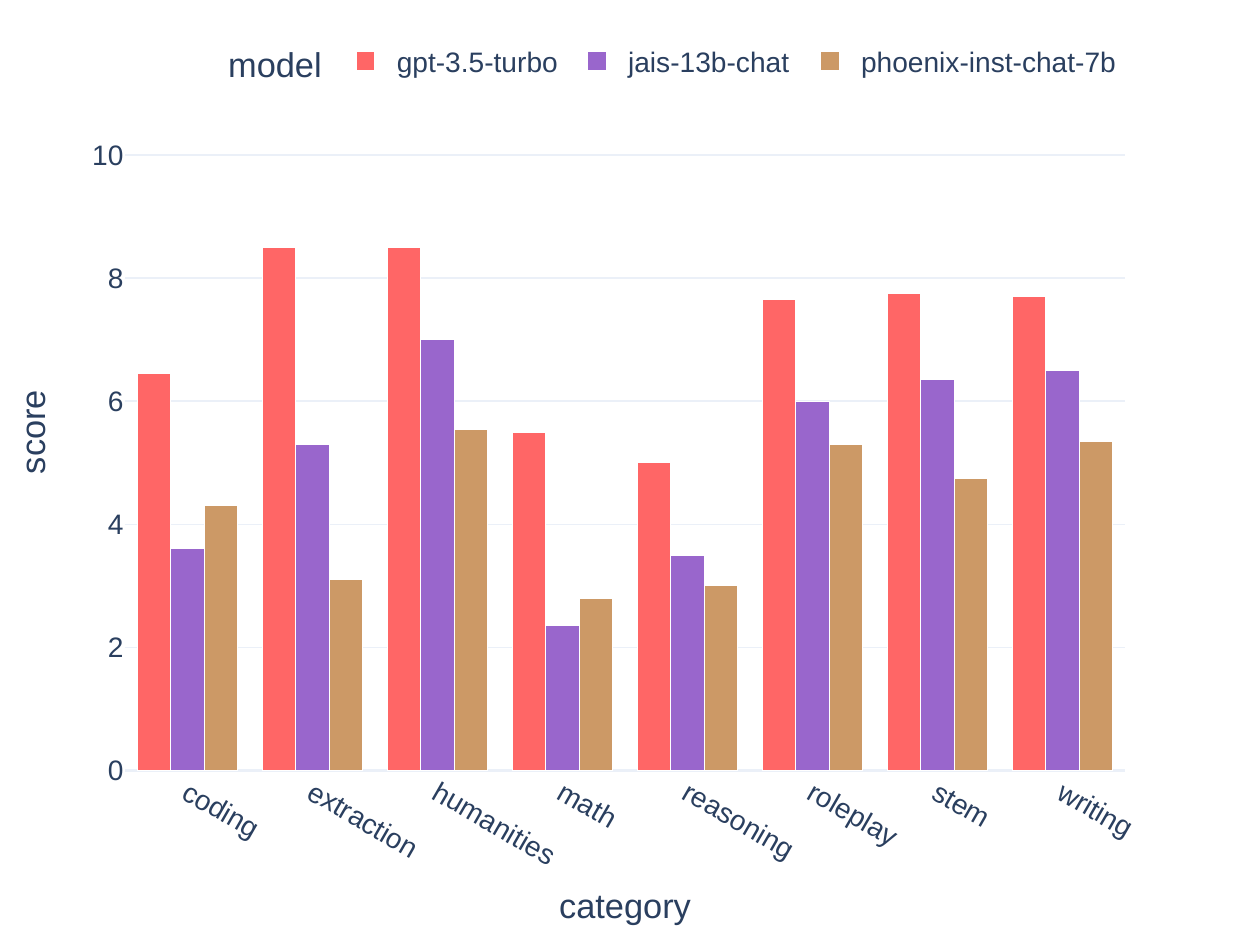}
    \caption{Average scores per category for three selected models evaluated on the \textsc{Arabic MT-Bench}.}
    \label{fig:3models}
\end{figure}

We note that the fine-tuning dataset of Jais-13B-chat is large with over 10M samples. The longer period needed for this fine-tuning  could raise additional challenges as it  
might increase the risk of catastrophic forgetting of knowledge gained during pre-training~\cite{luo2023empirical, he2021analyzing}.  For comparison,  Phoenix-inst-chat-7B is ranked second among the evaluated open models in our experiment. The model is fine-tuned from a BLOOMZ-7B1-MT base~\cite{phoenix-2023}. The fine-tuning dataset has 133 languages with 58\% English, 20.9\% Chinese and 0.8\% Arabic which is ranked 11th in language coverage, with a total of  267K instruction-tuning samples. The conversation-tuning dataset has 189K samples covering more than 40 languages. Despite its smaller size and wide coverage of languages, Phoenix-chat-7B achieves intriguing results. Figure~\ref{fig:3models} shows detailed comparison per category for Jais-13B-chat, Phoenix-inst-chat-7B and GPT-3.5. The two open LLMs had the lowest scores on math  and reasoning, whereas the highest scores are on roleplay, humanities and stem.

Vicuna-33B-v1.3 and MPT-30B-chat scored around 3 out 10, while they were not expected to have {any significant} skill in Arabic. One possible explanation is that given their size over 30B, they are able to maximize their multilingual skills effectively. This hypothesis needs further investigations. Despite their low performance, it is interesting to explore the model development in order to adapt for training multilingual LLMs.

\subsection{What is the effect of specifically targeting Arabic when training a model?}
\label{ssec:arabic_training}
\begin{figure}[ht]
    \centering
    \includegraphics[trim=0cm 0cm 0.5cm 0.5cm,clip,scale=0.5]{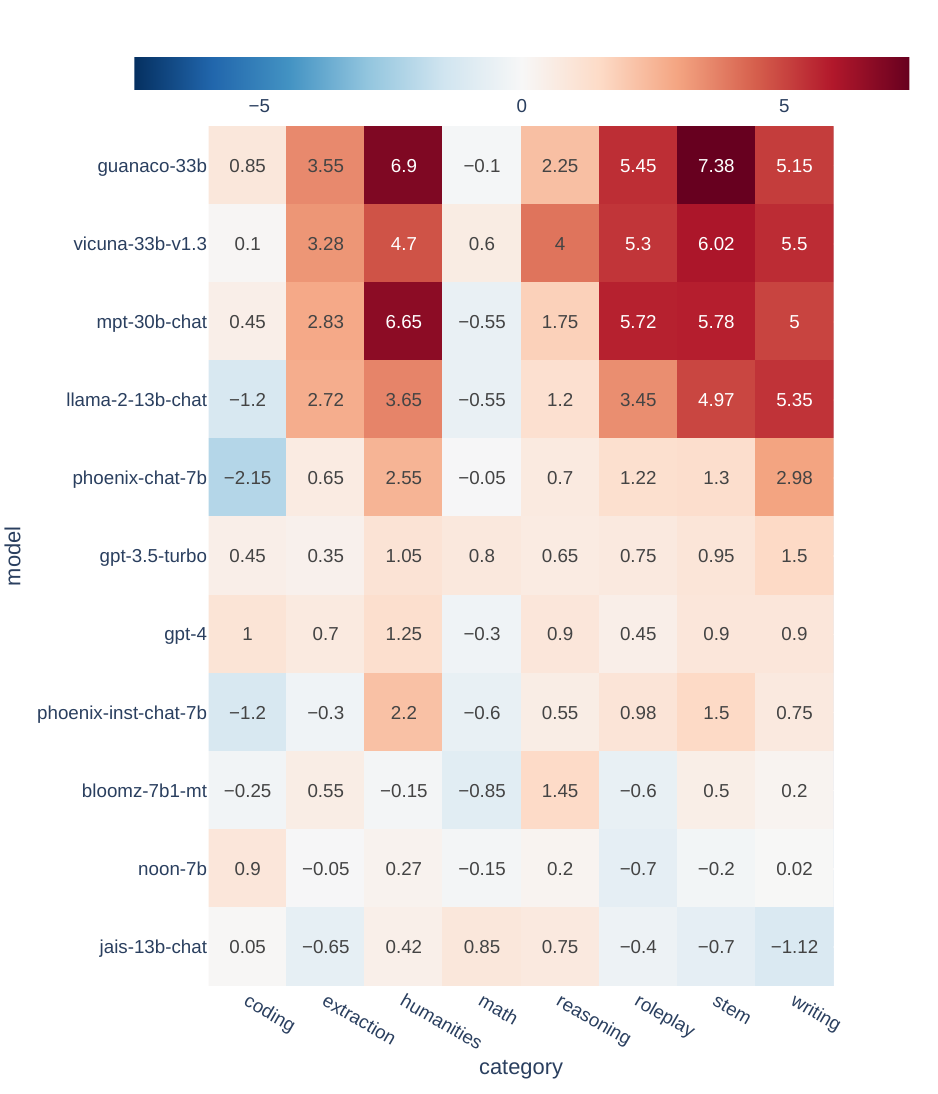}
    
    \caption{Difference of average MT scores between English and Arabic benchmarks per category. Positive values (red) indicate English answers are scored higher that the corresponding Arabic answers, while negative values (blue) indicate some advantage in Arabic. Neutral colors mean  a model is equally-competent in both languages.}
    \label{fig:diff}
\end{figure}

Figure~\ref{fig:diff} shows a heat map of the difference in score per category between the Arabic and the English benchmarks for the selected models. The models are sorted from top to bottom based on a decreasing score differences. {Warmer} cells in the figure indicate English advantage over Arabic for the same model and category, while {cooler} cells indicate Arabic advantage. 

The two GPT models reside in the neutral area, indicating comparable competency in  English and Arabic.  Not surprisingly, Models that have been pre-trained and fine-tuned on multilingual data (see Table~\ref{tab:models}) appear in the bottom half of the heat map, indicating some Arabic knowledge. {Also, it could be seen from the heatmap that coding and math are neutral, language-agnostic skills across models, as should be expected, while reasoning has a lingual side.}

Figure~\ref{fig:dotplot} shows the per-turn average scores of \textsc{Arabic MT-Bench}  on the X-axis and English MT-Bench  on the Y-axis 
for the selected models. Points closer to the diagonal line are models with similar average performance in Arabic and English, and the closer to the top right corner the better the model is on both languages. 
Most models are above the diagonal, and hence exhibit relatively superior skills in English compared to Arabic. This is likely due to the imbalance in the training and fine-tuning data between the two languages. Note that the LLaMa-based models are clustered together far from the diagonal, indicating lack in multilinguality, while BLOOMZ-7B1-MT and Noon-7B, both heavily multilingual, are on top of the diagonal.

\subsection{What is the effect of specifically fine-tuning on Arabic multi-turn instructions?}

\begin{figure}[ht]
    \centering
    \includegraphics[trim=0.8cm 1cm 1.8cm 0.5cm,clip,scale=0.6]{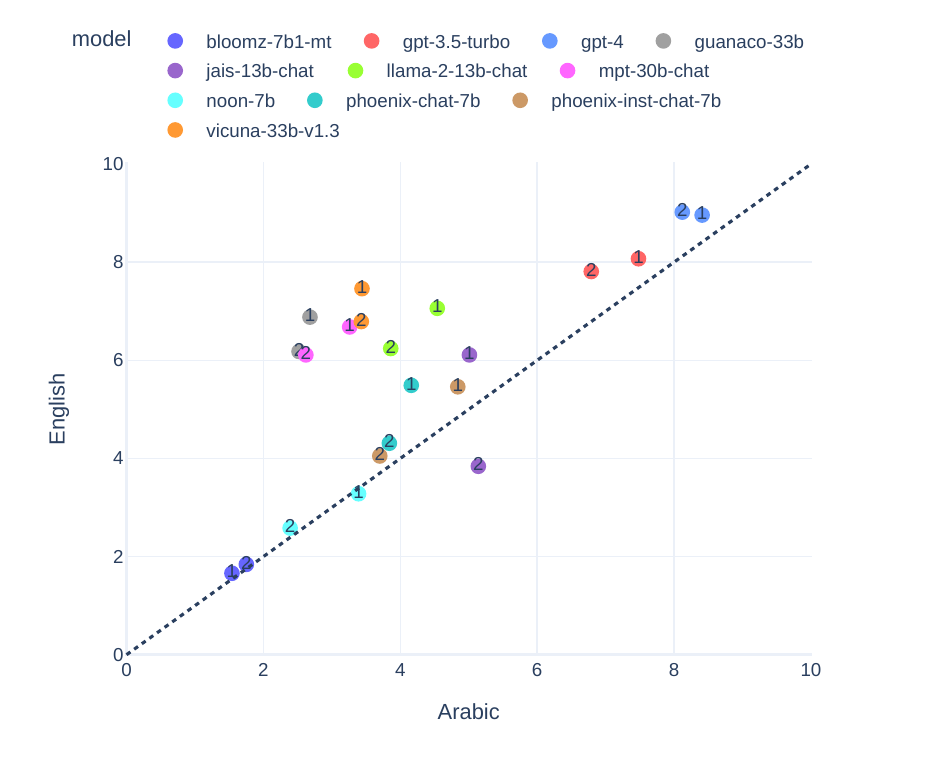}
    \caption{Scores in Arabic (X-axis) and English (Y-axis) MT-Bench for the first and second turn. The farther the model is from the diagonal, the bigger the gap in quality between the two languages. The farther Turn 2 is from Turn 1 for a model, the bigger the change in quality in responding to continued conversation.}
    \label{fig:dotplot}
\end{figure}

In Figure~\ref{fig:dotplot}, the two dots for each model represent the two turns, and their placement gives an insight into the ability of a model to engage in a conversation. Vertical drop between the two turns indicates diminished performance on English for the second turn, while horizontal shifts to the left indicates diminished performance on Arabic for the second turn. 

BLOOMZ-7B1-MT does not degrade on the second turn, even though it is not fine-tuned on conversational data~\cite{muennighoff2023crosslingual}, and it is the only model that is not affected in the second turn for both languages, while a capable model like GPT-4 had a slight improvement on the second turn for English but had a minor deterioration of the score for Arabic. 

On the other hand, Noon-7B has the largest drop in score between turns on Arabic. This model is built on top of BLOOM by instruct fine-tuning using a combination of  datasets with ColossalAI framework~\cite{bian2021colossal}. Noon-7B\footnote{\url{https://huggingface.co/Naseej/noon-7b}} used GPT-3.5-Turbo as a judge for evaluation on private data. 
%
We also observe that Jais-13B-chat has a large drop in English multi-turn instructions compared to a small drop in Arabic, {which might be caused by the ratio of Arabic to English instructions in its chat fine-tuning}.

Phoenix-chat-7B, Noon-7B and BLOOMZ-7B1-MT are all based on different variants of the backbone BLOOM-7B or BLOOMZ-7B. The resulting models vary a lot in terms of performance, indicating that a careful fine-tuning recipe is crucial for improving the capabilities of any base model.  

\subsection{How to select a good starting point LLM model to fine-tune  for Arabic instruction following?}

We consider the hypothetical optimal ensemble model defined by the maximum per-question  score across the  open models in our experiment. This characterizes an upper bound on the performance of any open LLMs ensemble made from these models. 
Based on our \textsc{Arabic MT-Bench}, the optimal ensemble model achieves an MT score of 6.70. This represents a 32\% increase in performance compared to the best individual open LLM (Jais, {5.08}).   Also it indicates that a collection of smaller models trained differently could capture various skills that might be difficult to capture together in one model without upping the model size. For the sake of contrast, for the English benchmark, the optimal ensemble model achieves a score of 8.2. 

\begin{figure}[hbt]
    \centering
    \includegraphics[trim=0.5cm 0.2cm 0.2cm 0.5cm,clip,scale=0.4]{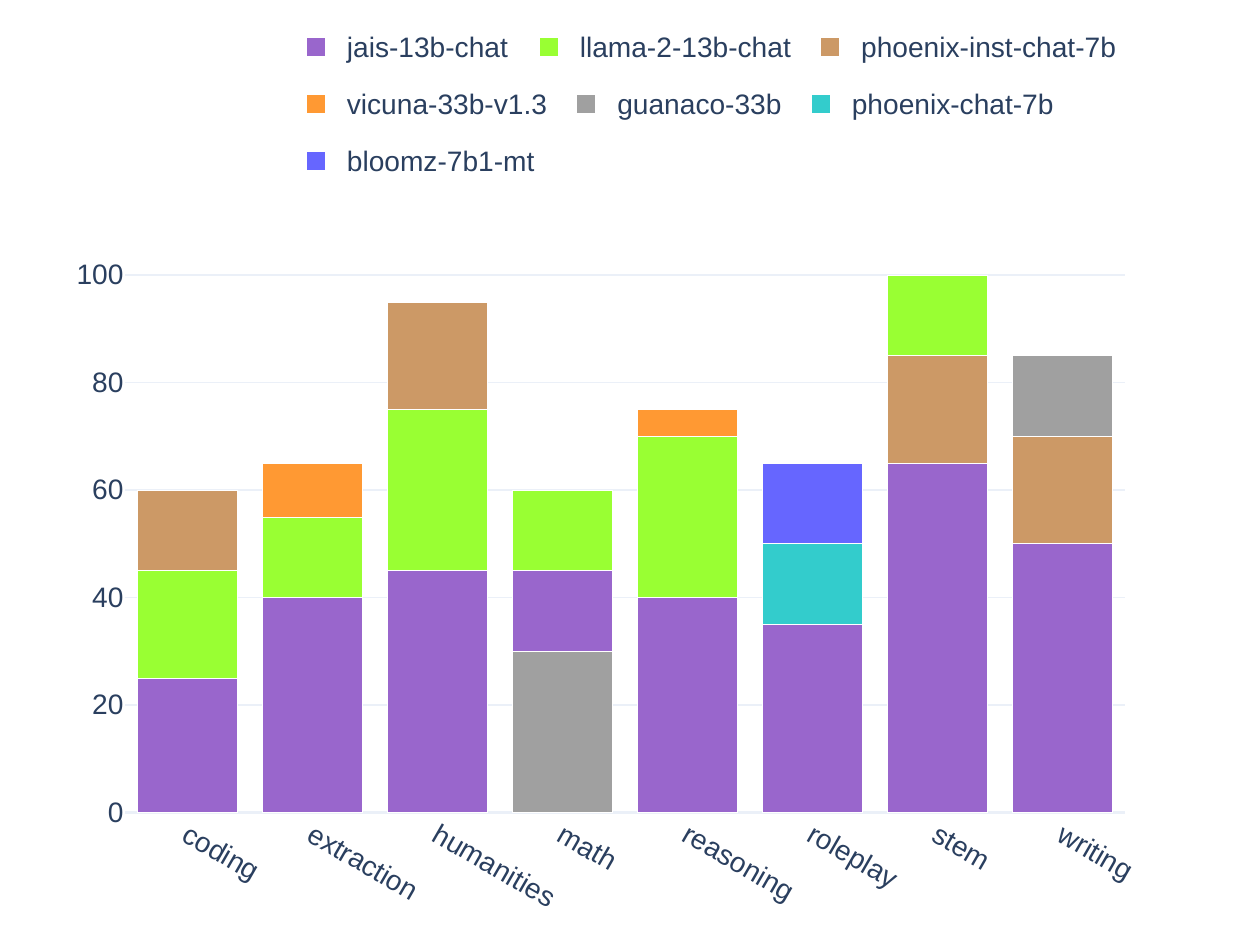}
    \caption{Contribution of the best three LLMs to the optimal ensemble model for each category. The Y-axis indicates how often a model was selected the best in terms of Arabic MT-score for the questions of a category. }
    \label{fig:ensemble}
\end{figure}

Figure~\ref{fig:ensemble} shows the contributions of the three highest-scoring  LLMs per category in the optimal Arabic ensemble model. 
We counted how often a model was  the best for a given  category 
and considered the top 3 models in each. Note that `best' here is relative to the performance of available LLMs, and is not an assessment of quality.

As the figure shows, Jais-13B-chat is the top model in five `literacy' categories, whereas math, coding and reasoning are shared with LLaMa-2-13B-chat, Guanaco-33B, and Phoenix-inst-chat-7b.  The challenge is how to define a criterion to select the best response among the ensemble LLMs. One possible approach is to ask each LLM to vote for the best answer and consider a majority vote, which will rely mainly on the ability of these small models to play the role of a judge in this limited context. We will leave investigating this  to future work. 


\section{Conclusion}
\label{sec:conclusion}
In this paper, we propose a framework for analyzing the effect of multilinguality on LLM performance in open-ended tasks. In particular, we  assessed the interaction between language, dialog and instruction following in Arabic and English for small open LLMs. We employ an LLM as a judge following the paradigm of MT-Bench. We show the effects of language on different categories of tasks and suggest ways to ensemble small LLMs to achieve better performance on the benchmark. 

In future work, we plan to extend the benchmark and analysis with more models and tasks, and investigate the viability of LLM ensemble models.

%




\section{Limitations}
We now discuss a number of limitations related to this study.
\subsection{Judging}
\begin{itemize}[leftmargin=*]
    \item  The use of an LLM as a judge for evaluating LLMs has issues related to bias. As reported in~\cite{zheng2023judging}, in pairwise comparisons, the judge tends to favor its own answers compared to other models. For example, that study shows that GPT-4 favors itself with 10\% higher win rate and Claude-V1 favors itself with 25\% higher win rate. On the other hand, GPT-3.5 does not appear to favor itself. 
    \item {Using GPT-4 as the judge and as an LLM under study might favor it in the scores. However, the score margin to the closet competitor is big enough to make any potential deviation in the scores insignificant, and we adhered to the original MT-Bench setup in the choice of judge in order to mirror the results and measure multilingual competency.}
    \item Other LLM judges than GPT-4 could be considered for evaluating the responses. However, the choice of alternative judges is currently rather limited when considering Arabic. The proficiency of models such as Claude or Bard in Arabic are not yet proven. Alternatively, multiple LLMs could be used for this task. A voting judgment mechanism could be considered over multiple open LLMs.
    \item While GPT-4 exhibits competence in Arabic, its proficiency in the language falls short of its mastery of English. This discrepancy may have had an impact on certain aspects of our analyses, especially when comparing Arabic results to English results.
    \item We used the same judgment prompt  as in the English MT-Bench for the purpose of consistency. However, we note that the judgment prompt does not acknowledge important aspects  such as safety and harmlessness of LLM responses. Also, the MT-score is a metric that combines multiple dimensions such as relevance, helpfulness, and creativity together to give an aggregate verdict. It might be useful to analyze  model performance separately on these dimensions for a better understanding. 
\end{itemize}
\subsection{Coverage}
\begin{itemize}[leftmargin=*]

    \item MT-Bench has a limited number of questions (160 in total considering both turns). This is likely not representative of the wide spectrum of tasks needed to effectively evaluate LLMs, and the authors of MT-Bench are acknowledging that by working to expand their benchmarking dataset to 1000 questions.  
    In addition, language-specific dimensions of conversation might require bespoke questions to test properly. 
    \item We only included a small number of models in the benchmark. During an initial screening, we excluded  several LLMs due to their limited capabilities in Arabic. We plan to extend our benchmark and include more LLMs in the future.
\end{itemize}


\bibliography{anthology,custom}
\bibliographystyle{acl_natbib}

\appendix

\section{Appendix}
\label{sec:appendix}

\subsection{English MT-Bench scores}
\label{appx:results}
Table~\ref{tab:leaderboard_en} shows the per-turn and average scores for the selected models on the original MT-Bench. 
 \begin{table}[!h]
    \centering
    
\begin{tabular}{l|l|l|r}
\toprule
\textbf{Model} & \textbf{Turn1} &  \textbf{Turn2} & \textbf{Avg}\\
\midrule
\textit{GPT-4} & 8.96 & 9.02 & 8.99 \\
\textit{GPT-3.5-turbo} & 8.07 & 7.81 & 7.94 \\
\hline
\textit{Vicuna-33B-v1.3} & 7.46 & 6.79 & 7.12 \\
\textit{Llama-2-13B-chat} & 7.06 & 6.24 & 6.65 \\
\textit{Guanaco-33B} & 6.88 & 6.18 & 6.53 \\
\textit{Tulu-30B} & 7.02 & 5.85 & 6.43 \\
\textit{MPT-30B-chat} & 6.68 & 6.11 & 6.39 \\
\textit{Jais-13B-chat} & 6.11 & 3.84 & 4.97 \\
\textit{Phoenix-chat-7B} & 5.49 & 4.31 & 4.90 \\
\textit{Phoenix-inst-chat-7B} & 5.46 & 4.05 & 4.75 \\
\textit{Noon-7B} & 3.28 & 2.58 & 2.93 \\
\textit{Bloomz-7B1-mt} & 1.66 & 1.84 & 1.75 \\
\textit{Bloomz-7B1} & 1.39 & 1.85 & 1.62 \\
\bottomrule
\end{tabular}
    \caption{Results of benchmarked LLMs on English \textsc{MT-Bench} (scores between 0-10). showing for each model average scores per turn,  and average score across all questions and turns.}
    \label{tab:leaderboard_en}
\end{table}

\subsection{Prompts for LLM Judge}
Figure~\ref{fig:prompt1} shows the judging prompt for the first-turn questions in MT-Bench, and Figure~\ref{fig:prompt2} shows the prompt for the second-trun questions.
 \begin{figure}[h!]
    \centering
    \begin{tikzpicture}
        \node[draw, rounded corners, align=center] {
            \begin{minipage}{\columnwidth}
                   \small \texttt{Please act as an impartial judge and evaluate the quality of the response provided by an AI assistant to the user question displayed below. \sethlcolor{mygreen}\hl{Your evaluation should consider factors such as the helpfulness, relevance, accuracy, depth, creativity, and level of detail of the response}. Begin your evaluation by providing a short explanation. Be as objective as possible. After providing your explanation, you must rate the response on a scale of 1 to 10 by strictly following this format: [[rating]], for example: "Rating: [[5]]".} 
            \end{minipage}
        };
    \end{tikzpicture}
    \caption{LLM judge  first turn's prompt. The highlighted text indicates the evaluation criteria. }
    \label{fig:prompt1}
\end{figure}

\begin{figure}[h!]
    \centering
    \begin{tikzpicture}
        \node[draw, rounded corners, align=center] {
            \begin{minipage}{\columnwidth}
                   \small \texttt{Please act as an impartial judge and evaluate the quality of the response provided by an AI assistant to the user question displayed below. \sethlcolor{mygreen}\hl{Your evaluation should consider factors such as the helpfulness, relevance, accuracy, depth, creativity, and level of detail of the response}. \sethlcolor{myorange}\hl{You evaluation should focus on the assistant's  answer to the second user question}. Begin your evaluation by providing a short explanation. Be as objective as possible. After providing your explanation, you must rate the response on a scale of 1 to 10 by strictly following this format: [[rating]], for example: "Rating: [[5]]".} 
            \end{minipage}
        };
    \end{tikzpicture}
    \caption{LLM judge  second turn's prompt. The highlighted text in green indicates the evaluation criteria. The highlighted text in orange indicates the instruction to focus the evaluation on the answer of the second question.}
    \label{fig:prompt2}
\end{figure}

\subsection{Chosen Models}
\label{sec:chosen_models}
\begin{itemize}[leftmargin=*]
\item GPT-4: a proprietary  multilingual chatbot by OpenAI, trained on public and proprietary data and fine-tuned using reinforcement learning with human and AI-generated feedback. Allows 8k and 32k prompts~\cite{openai2023gpt4}.

\item GPT-3.5-turbo: the predecessor of GPT-4 with 175B parameters.

\item Jais-13B-chat: A 13B parameter model that follows the GPT-3 architecture, pre-trained on  279B English and 116B Arabic tokens, then fine-tuned on 5.9 million English and 3.8 million Arabic  supervised multi-turn instructions, and further fine-tuned for safety~\cite{sengupta2023jais}.

\item Phoenix-chat-7B: A BLOOMZ-based 7B parameter model fine-tuned for dialog  using online ChatGPT records and multi-round conversations
~\cite{phoenix-2023}.

\item Phoenix-inst-chat-7B: Another  7B model from the Phoenix family, fine-tuned not only for conversations  but also for multilingual instruction following using self-instruct and translators.
\item Vicuna-33B-v1.3: A 33B LLaMa-based model, fine-tuned on a ShareGPT.com dataset for instruction following and multi-turn dialog~\cite{zheng2023judging}.

\item MPT-30B-Chat: A fine-tuned version of MPT-30B which is an encoder-only transformer model trained on 1T English tokens. MPT-30B-Chat was fine-tuned for chat on a number of public datasets including ShareGPT-Vicuna, Camel-AI, GPTeacher, Guanaco and Baize~\cite{MosaicML2023Introducing}.

\item Noon-7B: A BLOOM-based 7B parameter model, fine-tuned on 110k Arabic instructions from translated datasets including  GPT-4 responses to Alpaca quesitons, Dolly, TruthfulQA, Grade School Math in addititon to self-instruct questions in Arabic.

\item Guanaco-33B: A LLaMa-based model with 33B parameters, 
fine-tuned on 534k multiligual instructions using the OASST1 dataset. Not chat trained~\cite{dettmers2023qlora}.
\item PolyLM-13B: A decoder-only model of 13B parameters, pre-trained on a multilingual training data of 640B tokens, and fine-tuned on MULTIALPACA that contains 132K  multilingual instructions generated in a self-instruct fashion.~\cite{wei2023polylm} 
\item Llama-2-13B-Chat: A  member of Llama2 auto-regressive transformer models with 13B parameters, pre-trained on 2T tokens with 4k context, and fine-tuned for multi-turn dialog using supervised fine-tuning  on public instruction datasets and reinforcement learning with human feedback over more than 1 million human annotations~\cite{touvron2023llama}.
\item BLOOMZ-7B1: A multilingual decoder-only transformer model trained on 350B tokens including 45 natural languages, and fine-tuned on xP3, a multitask and multilingual instruction dataset. Recommended for prompting in English.~\cite{muennighoff2023crosslingual}
\item BLOOMZ-7B1-MT: A version of BLOOMZ-7B1 fine-tuned on xP3mt, a multitask and multilingual instruction dataset with machine-translated prompts in 20 languages. Recommended for prompting in non-English.
\item Tulu-30B: A LLaMa-based 33B model fine-tuned on  number of publicly-available instruction datasets including FLAN V2, CoT, Dolly, Open Assistant 1, GPT4-Alpaca, Code-Alpaca, and ShareGPT.~\cite{wang2023far}
\end{itemize}

\subsection{Arabic questions and reference answers}
\label{sec:appx_questions}
The full set of questions and reference answers of \textsc{Arabic MT-Bench} are {available at \url{https://huggingface.co/spaces/QCRI/mt-bench-ar}}.

Here in Table~\ref{tbl:example_curated} we present a sample of the curated questions.

\begin{table*}[!t]
\begin{tabular}{p{0.03\columnwidth}|p{0.05\columnwidth}|p{1.8\columnwidth}}
                  \hline
\multirow{7}{*}{\rot{Writing}}& \multirow{4}{*}{T1} &   \begin{arabtext}\small اكتب فقرة تصف فيها سوقاً مزدحماً وضمن فيها تفاصيل حسية كالروائح والأصوات والعناصر المرئية لخلق تجربة غامرة للقارئ. الرجاء الإجابة باللغة العربية\end{arabtext}  \\ \cline{2-3}
& \multirow{3}{*}{T2} & \begin{arabtext}\small
أعد صياغة إجابتك السابقة مستهلاً كل جملة بالحرف الأبجدي التالي للجملة التي قبلها بدءًا من الحرف ب. الرجاء الإجابة باللغة العربية
                          \end{arabtext}
\\ \Xhline{2\arrayrulewidth}

\multirow{11}{*}{\rot{Roleplay}}& \multirow{7}{*}{T1} & \begin{arabtext}\small
    الرجاء تمثل دور مترجم إلى اللغة العربية مكلف بتصحيح الإملاء وتحسين اللغة. بغض النظر عن اللغة التي أستخدمها في السؤال عليك تحديدها وترجمتها بلغة عربية رشيقة. استخدم تعبيرات بليغة وعالية وحافظ على المعنى الأصلي للجملة. ركز على تقديم التصحيحات والتحسينات فقط. جملتي الأولى هي \\ <`When the going gets tough, the tough get going'>\\الرجاء الإجابة باللغة العربية
\end{arabtext}\\ \cline{2-3}
& \multirow{4}{*}{T2} &  \begin{arabtext}\small
 <`Ich verstehe nichts'>\\
الرجاء الإجابة باللغة العربية
\end{arabtext}\\ \Xhline{2\arrayrulewidth}

\multirow{8}{*}{\rot{Roleplay}}     & \multirow{5}{*}{T1} &  \begin{arabtext}\small
جسد شخصية علاء الدين من `علاء الدين والمصباح السحري' طوال هذه المحادثة. لا تقل `بصفتي علاء الدين' في بداية الجمل. سؤالنا الأول هو: ما هو الشيء المفضل لديك في كونك علاء الدين؟\\
الرجاء الإجابة باللغة العربية

\end{arabtext}  \\ \cline{2-3}
& \multirow{4}{*}{T2} & \begin{arabtext}
    \small ما رأيك في <GPT-4> كبديل عن جني المصباح؟\\
الرجاء الإجابة باللغة العربية
\end{arabtext}  \\ \Xhline{2\arrayrulewidth}

\multirow{6}{*}{\rot{Reasoning}}& \multirow{2}{*}{T1} & \begin{arabtext}\small  لداود ثلاث أخوات، لكل واحدة منهن أخ واحد. كم أخاً لداود؟
  الرجاء الإجابة باللغة العربية
\end{arabtext}\\ \cline{2-3}
& \multirow{3}{*}{T2} &  \begin{arabtext}\small
إذا غيرنا السؤال السابق وافترضنا أن كل أخت لداود لها أخوان اثنان، فكم سيكون عدد إخوة داود؟
الرجاء الإجابة باللغة العربية
\end{arabtext}\\ \Xhline{2\arrayrulewidth}
\end{tabular}
\caption{A sample of translated and curated questions from {\sc{Arabic MT-Bench}} in categories Writing, Roleplay and Reasoning. T1 and T2 denote the first and second turn (follow-up) questions, respectively.}
\label{tbl:example_curated}
\end{table*}



\end{document}